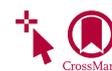

# Development and validation of an interpretable machine learning-based calculator for predicting 5-year weight trajectories after bariatric surgery: a multinational retrospective cohort SOPHIA study

*Patrick Saux\*, Pierre Bauvin\*, Violeta Raverdy, Julien Teigny, Hélène Verkindt, Tomy Soumphonphakdy, Maxence Debert, Anne Jacobs, Daan Jacobs, Valerie Monpellier, Phong Ching Lee, Chin Hong Lim, Johanna C Andersson-Assarsson, Lena Carlsson, Per-Arne Svensson, Florence Galtier, Guelareh Dezfoulian, Mihaela Moldovanu, Severine Andrieux, Julien Couster, Marie Lepage, Erminia Lembo, Ornella Verrastro, Maud Robert, Paulina Salminen, Geltrude Mingrone, Ralph Peterli, Ricardo V Cohen, Carlos Zerrweck, David Nocca, Carel W Le Roux, Robert Caiazzo, Philippe Preux, François Pattou*

## Summary

**Background** Weight loss trajectories after bariatric surgery vary widely between individuals, and predicting weight loss before the operation remains challenging. We aimed to develop a model using machine learning to provide individual preoperative prediction of 5-year weight loss trajectories after surgery.

**Methods** In this multinational retrospective observational study we enrolled adult participants (aged ≥18 years) from ten prospective cohorts (including ABOS [NCT01129297], BAREVAL [NCT02310178], the Swedish Obese Subjects study, and a large cohort from the Dutch Obesity Clinic [Nederlandse Obesitas Kliniek]) and two randomised trials (SleevePass [NCT00793143] and SM-BOSS [NCT00356213]) in Europe, the Americas, and Asia, with a 5 year follow-up after Roux-en-Y gastric bypass, sleeve gastrectomy, or gastric band. Patients with a previous history of bariatric surgery or large delays between scheduled and actual visits were excluded. The training cohort comprised patients from two centres in France (ABOS and BAREVAL). The primary outcome was BMI at 5 years. A model was developed using least absolute shrinkage and selection operator to select variables and the classification and regression trees algorithm to build interpretable regression trees. The performances of the model were assessed through the median absolute deviation (MAD) and root mean squared error (RMSE) of BMI.

**Findings** 10 231 patients from 12 centres in ten countries were included in the analysis, corresponding to 30 602 patient-years. Among participants in all 12 cohorts, 7701 (75·3%) were female, 2530 (24·7%) were male. Among 434 baseline attributes available in the training cohort, seven variables were selected: height, weight, intervention type, age, diabetes status, diabetes duration, and smoking status. At 5 years, across external testing cohorts the overall mean MAD BMI was 2·8 kg/m² (95% CI 2·6–3·0) and mean RMSE BMI was 4·7 kg/m² (4·4–5·0), and the mean difference between predicted and observed BMI was −0·3 kg/m² (SD 4·7). This model is incorporated in an easy to use and interpretable web-based prediction tool to help inform clinical decision before surgery.

**Interpretation** We developed a machine learning-based model, which is internationally validated, for predicting individual 5-year weight loss trajectories after three common bariatric interventions.

**Funding** SOPHIA Innovative Medicines Initiative 2 Joint Undertaking, supported by the EU's Horizon 2020 research and innovation programme, the European Federation of Pharmaceutical Industries and Associations, Type 1 Diabetes Exchange, and the Juvenile Diabetes Research Foundation and Obesity Action Coalition; Métropole Européenne de Lille; Agence Nationale de la Recherche; Institut national de recherche en sciences et technologies du numérique through the Artificial Intelligence chair Apprenf; Université de Lille Nord Europe's I-SITE EXPAND as part of the Bandits For Health project; Laboratoire d'excellence European Genomic Institute for Diabetes; Soutien aux Travaux Interdisciplinaires, Multi-établissements et Exploratoires programme by Conseil Régional Hauts-de-France (volet partenarial phase 2, project PERSO-SURG).

## Introduction

The prevalence of obesity has increased globally over the past two decades.[1] Obesity is a heterogeneous condition associated with several complications and metabolic manifestations across individuals, ultimately increasing the risk of all-cause mortality.[2] Bariatric surgery, although not first-line therapy, has emerged as an effective treatment for sustained weight loss,[3] (Prof D Nocca MD PhD), **CHU de Montpellier, University of Montpellier, Montpellier, France; Clinical Investigation Center 1411, INSERM, CHU de Montpellier, University of Montpellier, Montpellier, France** (F Galtier); **Centre Hospitalier Valenciennes, Valenciennes, France** (G Dezfoulian MD, M Moldovanu MD); **Centre Hospitalier Arras, Arras, France** (S Andrieux MD); **Centre Hospitalier Boulogne-sur-Mer, Boulogne-sur-Mer, France** (J Couster MD, M Lepage MD); **Fondazione Policlinico Universitario A Gemelli IRCCS, Rome, Italy** (E Lembo MD, O Verrastro PhD, Prof G Mingrone MD PhD); **Università Cattolica del Sacro Cuore Rome, Rome, Italy** (E Lembo, O Verrastro, Prof G Mingrone); **Department of Digestive Surgery, Center of Bariatric Surgery, Hopital Edouard Herriot, Hospices Civils de Lyon, Lyon, France** (Prof M Robert MD PhD); **Division of Digestive Surgery and Urology, Turku University Hospital, Turku, Finland** (Prof P Salminen MD PhD); **Department of Surgery, University of Turku, Turku, Finland** (Prof P Salminen); **University of Basle, Basle, Switzerland** (R Peterli MD); **Clarunis, Department of Visceral Surgery, University Centre for Gastrointestinal and Liver Diseases, St Clara Hospital and University Hospital Basle, Basle, Switzerland** (R Peterli); **The Center for Obesity and Diabetes, Oswaldo Cruz German Hospital, São Paulo, Brazil** (R V Cohen MD PhD); **Clínica Integral de Cirugía para la Obesidad y Enfermedades Metabólicas, Hospital General Tláhuac, Mexico City, Mexico** (Prof C Zerrweck MD); **University College Dublin, Dublin, Ireland** (Prof C W Le Roux MD PhD)

Correspondence to:
Prof Philippe Preux, Université de Lille, CNRS, Inria, Centrale Lille, UMR 9189 – CRIStAL, F-59000 Lille, France
**philippe.preux@inria.fr**
or
Prof François Pattou, Université de Lille, Inserm, CHU Lille, Institut Pasteur de Lille, U1190 - EGID, F-59000, Lille, France
**francois.pattou@univ-lille.fr**

### Research in context

**Evidence before this study**
Obesity is a heterogeneous condition that increases the risk of all-cause mortality. Bariatric surgery, although not first-line therapy, is an effective treatment for sustained weight loss, improving obesity-related complications and life expectancy. However, despite comprehensive preoperative assessment of each candidate, long-term weight loss outcomes are heterogeneous as regards to changes over time, differences between procedures, and between individuals.
We searched PubMed, Embase, and the Cochrane Library using the terms "bariatric surgery", "postoperative weight loss", "weight loss prediction", and "prediction model" for studies that investigated models of weight loss after Roux-en-Y gastric bypass (RYGB), sleeve gastrectomy, and adjusted gastric banding (AGB) and used a prospective or retrospective design published from database inception up until Jan 21, 2021. We included English and French language studies. Most previous studies were restricted to the early postoperative period or undermined by the high proportion of patients lost to follow-up, or both. Some prediction models also incorporated early weight loss (before 6 months), to predict longer outcomes (beyond 2 years), and can therefore not be used before the operation. Previous attempts to predict longer-term weight loss after bariatric surgery have used multivariable regressions. However, such methods assume that the relationship between the dependent and the independent variable is linear, which might not always be the case.

**Added value of this study**
In the present study, we developed a machine learning model that provides accurate individual weight trajectories expected during 5 years after bariatric surgery, based on seven simple preoperative variables: age, weight, height, smoking history, type 2 diabetes status and duration, and the type of intervention. These variables are readily available in a variety of clinical settings without interpretation and do not require laboratory tests. The model was incorporated into an easy-to-use and interpretable web-based tool. This study is the first to provide preoperative predictions of weight trajectories up to 5 years after surgery based on machine learning, simultaneously for three of the most common types of surgery (RYGB, sleeve gastrectomy, and AGB). The present study also showed the effect of diabetes duration and smoking, which were not previously included in weight loss surgery prediction models.

**Implications of all the available evidence**
This model could help to refine individual weight loss trajectory prediction in routine clinical practice, by being an accurate and simple strategy to inform clinical decisions for both health-care providers and patients before surgery, enabling precision medicine and individualised patient management.

resulting in long-term improvement in obesity-related complications,[4] and prolonged life expectancy.[5] Despite a comprehensive preoperative assessment of each patient, it is challenging to forecast weight loss outcomes following the intervention. Indeed, weight loss changes over time and varies between procedures and between individuals.[6,7]

A reasonable estimation of the expected weight loss trajectory after bariatric surgery would help inform clinical decisions by patients and health-care providers. Therefore, multiple models have been proposed to predict postoperative weight loss.[8] For such models to be clinically relevant, they need to predict at least 5 years of weight loss outcome.[4] However, most previous studies were restricted to the early postoperative period[8] or undermined by the high proportion of patients lost to follow-up, or both.[9] Some prediction models also incorporated early weight loss (before 6 months), to predict longer outcomes (beyond 2 years),[10,11] and can therefore not be used before the operation.

Previous attempts to predict longer-term weight loss after bariatric surgery have used multivariable regressions.[8] However, such methods assume that the relationship between the dependent and the independent variable is linear, which might not always be the case. Additionally, multivariable regressions might lead to difficulties in accounting for interactions.[12]

Moreover, although coefficients in linear regression and the odds in logistic regression are relatively easy to understand, they are not easy to apply in clinical decision making. In contrast, machine learning methods have the potential to distinguish subtle, non-linear patterns in data that are often not accessible using traditional approaches such as logistic regression.[13] Likewise, machine learning models have outperformed logistic regression in preoperative risk stratification using National Surgical Quality Improvement Program data.[14] Currently only a few studies have applied artificial intelligence (neural network) to predict early weight loss after bariatric surgery, and none of them were externally validated.[15]

Therefore, the aims of the present study were (1) to use machine learning to develop a system predicting post-operative weight loss trajectory, using information gathered by protocol-driven, comprehensive preoperative assessment and repeated postoperative weight assessments in a large prospective cohort study of patients submitted to bariatric surgery; (2) to validate the performance of the proposed model globally, using multiple external prospective cohorts and randomised controlled trials; and (3) to incorporate the results into an easy-to-use and interpretable web-based tool providing individual preoperative prediction of postoperative weight loss trajectory.

## Methods

### Study design and participants

In this multinational retrospective observational study, we used data from ten cohorts of adult patients submitted for the first time to Roux-en-Y gastric bypass (RYGB), sleeve gastrectomy, and adjusted gastric banding (AGB), from eight countries. All patients had up to 5 years of postoperative data available and were aged 18 years or older. We excluded patients with a previous history of bariatric surgery, because the preoperative weights measured before reintervention already accounted for the effect of past interventions, which would have added a bias in the calculated weight loss outcomes. We also excluded patients with large delays between scheduled and actual visits related to postoperative complications. In case of missing follow-up visits, patients were kept in the analysis but censored at the corresponding dates. Patients not expected at a given time (recent interventions) were also censored after the last completed visit (appendix p 3).

The training cohort consisted of patients who were prospectively enrolled at the time of primary bariatric surgery in two longitudinal cohort studies evaluating the long-term outcome of bariatric surgery: Atlas Biologique de l'Obésité Sévère (ABOS; NCT01129297) in Lille, France, between Feb 10, 2006, and Nov 2, 2020, and BAREVAL (NCT02310178) in Montpellier, France, between April 8, 2014, and April 28, 2020.

The prediction model was validated, using eight external testing cohorts from France (Projet régional de Recherche Clinique en Obésité Sévère [PRECOS; NCT03517072] and Lyon [NCT02139813]), the Netherlands (the Dutch Obesity Clinic, Nederlandse Obesitas Kliniek [NOK]),[10] Sweden (the Swedish Obese Subjects [SOS] study),[3] Italy (NCT01581801 and NCT00888836),[16] Singapore (Singapore General Hospital [SGH]),[17] Brazil (Center for the treatment of Obesity and Diabetes [COD], Hospital Oswaldo Cruz, Sao Paulo, Brazil),[18] and Mexico.[19]

Additional external validation was conducted in participants of two registered and previously published randomised, open-label, multicentre trials which compared patients submitted to RYGB with sleeve gastrectomy in Finland (SleevePass, NCT00793143)[20] and Switzerland (Swiss Multicenter Bypass or Sleeve Study [SM-BOSS], NCT00356213).[7] The individual-level 5 year-data of these two studies have been merged into a single analysis.[21]

Details about study protocols and data collection for these two cohorts are in the appendix (p 7). Study participants self-reported sex data and were provided with two options (male or female).

Study participants in all cohorts gave written informed consent. All centres obtained ethics approval for their respective studies.

### Outcomes

The primary study outcome was the prediction of BMI at 5 years after bariatric surgery. Secondary outcomes were weight loss at earlier postoperative visits (at months 1, 3, 12, and 24), expressed as weight (kg), percent of total weight loss (TWL), calculated as:

$$TWL = (visit\ weight - preoperative\ weight)/preoperative\ weight \times 100$$

and percent of excess weight loss (EWL) calculated as:

$$EWL = (preoperative\ BMI - visit\ BMI)/(preoperative\ BMI - 25) \times 100$$

### Model development

To derive the model, the training cohort was divided into two subsets: a training subset consisting of 80% of randomly selected patients, and an internal testing subset consisting of 20% of patients.

We first performed preprocessing of all patients' baseline characteristics. Because the ABOS cohort had many preoperative attributes per patient (appendix pp 40–60), we ran a feature selection algorithm on this patient subgroup to extract the most statistically relevant ones concerning outcome prediction using the Least Absolute Shrinkage and Selection Operator (LASSO).[22]

To develop the model, first we further leveraged a class of machine learning algorithms called decision trees to learn meaningful subgroups of patients that share statistical similarities in their baseline characteristics, and second, to fit a TWL prediction model for each subgroup. For instance, decision trees can predict weight loss when they are trained on a heterogeneous cohort of different bariatric interventions such as RYGB, sleeve gastrectomy, and AGB, according to the type of intervention as well as using other variables such as the age at intervention, BMI, and other clinical features.

To calibrate the decision trees, we used LASSO-extracted features as input for the classification and regression trees (CART) algorithm.[23] A workflow diagram of the machine learning process is in the appendix (p 24). The algorithm was calibrated on the training subset of the training cohort. We further compared the predicted TWL to the observed outcomes of patients in the testing subset of the training cohort (internal validation). Details of the model development are in the appendix (p 8).

Additionally, we compared that approach with other methods: classic linear models, linear mixed effect model, random forest model, and CART on all variables without LASSO using instead pruning for feature selection (appendix p 9).

### Model validation in external cohorts

The model was externally validated by comparing the observed (TWL$_i$) and predicted total weight loss:

$$\widehat{TWL}_i$$

at each visit for each participant ($i$) of eight distinct testing cohorts (NOK, SGH, SOS, PRECOS, Roma, Lyon,

COD, and Mexico). For better readability, weight loss was also computed as BMI by converting predicted TWL into predicted weights:

$$\text{predicted weight} = \text{preoperative weight} \times (1 - \widehat{TWL}_i / 100)$$

The performance of the prediction model was calculated at each visit date, and expressed by using the standard metric median absolute deviation (MAD):

$$\text{MAD} = \text{median of } |TWL_i - \widehat{TWL}_i|$$

which measures dispersion of predicted TWL around the true values while being robust to outliers. We also calculated root mean squared error (RMSE):

$$\text{RMSE} = \text{square root of the mean of } (TWL_i - \widehat{TWL}_i)^2$$

which jointly measures the model prediction bias and variance, but is more sensitive to outliers than MAD because the square amplifies them. These two indices were also expressed as normalised ratios as percentage of observed BMI for each visit.

We used Bland–Altman plots with actual versus predicted BMI at each specific timepoint (month 12, month 24, and month 60) to assess model calibration.

### Model validation in randomised controlled trials

Additionally, we used the individual data from the two randomised clinical trials (SLEEVEPASS[24] and SM-BOSS[7]) to replicate with our model the previously reported comparison of RYGB versus sleeve gastrectomy in terms of weight loss.[21] The original report combined and analysed weight follow-up data from the two studies for up to 5 years using a linear mixed model.[21] In our study, we replaced the observed individual weight loss values with those predicted by our machine learning model at each timepoint by our machine learning model and analysed them using the same linear mixed model described in the original report to compare the predicted and observed mean (95% CI) difference in weight loss between the two operations.

### Statistical analysis

Patients' characteristics were reported for each cohort as mean (SD) for continuous traits and n (%) for categorical variables. Comparison of median weight loss between two groups was performed using the Mann-Whitney $U$ test, and between three or more groups using Kruskal-Wallis one-way ANOVA. MAD and RMSE were displayed as their estimates and 95% CIs, estimated by bootstrap (bias-corrected and accelerated method, n=10 000 replications). Weight loss and BMI median trajectories of participants submitted to each operation in each cohort were illustrated as a function of time using a non-linear smoothing of the values observed at discrete times (appendix p 4), which is not part of the validation and performance assessment. The trajectories of patients are displayed along with prediction intervals of predicted BMI, calculated as prediction plus 25th percentile of error and prediction plus 75th percentile of error.

Patient features containing more than 50% of missing values were excluded from the analysis. The remaining missing values were handled in two ways. For the LASSO analysis, the predictive mean matching method was used based on key characteristics at baseline: weight, sex, age, operation type, and presence of type 2 diabetes and its duration.[25] We imputed n=10 sets of data, and selected variables by pooling the variable selected by LASSO in each dataset. The decision tree algorithm uses surrogate variables for the CART analysis in the case of missing data.[26]

The analysis was performed using R software version 3.6.3, the rpart library for the CART implementation, and the glmnet and glinternet libraries for the LASSO implementation. An online tool was developed based on two components: a front-end graphical user interface coded with JavaScript using the React library, and a back-end prediction and smoothing model coded in Python using the Flask microweb framework. We used the TRIPOD AI guidelines[27] to report the prediction model's development and validation (appendix pp 61–63).

For comparison, we also analysed results from previously published postoperative weight loss prediction models[8] in the appendix (p 10).

### Role of the funding source

The funders of the study had no role in study design, data collection, data analysis, data interpretation, or writing of the report.

## Results

The training cohort from France (n=1493), and the testing cohorts from Europe (n=7137), the Americas (n=167), Asia (n=977), and the two randomised controlled trials used for validation (n=457), represented a total of 10 231 patients from 12 centres in ten countries, corresponding to 30 602 patient-years. The baseline characteristics of participants of each cohort and the proportion of each type of operation performed are in table 1. The overall trajectories of the median (IQR) BMI and TWL observed during the 5 years after each operation are shown in figure 1, and in the appendix (pp 12–23) for the various training and testing cohorts. Individual weights at baseline ranged from 65 kg to 295 kg (SD 25·6) and BMI from 26·7 kg/m² to 94·1 kg/m² (SD 7·5). Age at intervention ranged from 18 to 74 years. Among the 10 231 participants in all 12 cohorts, 7701 (75·3%) were female, 2530 (24·7%) were male, and 2882 (28·2%) had type 2 diabetes at baseline. RYGB was the most frequent operation (6691 [65·4%] of 10 231), followed by sleeve gastrectomy (2872 [28·1%]), and AGB (668 [6·5%]). At 5 years, the median TWL was 26·8% (IQR 19·8–34·0), ranging from –13·3% to 62·7%. Overall, the general shapes of weight loss trajectories for each operation were similar among cohorts, with a nadir weight loss reached between 1 and 2 years, followed by limited weight regain afterwards: median 18·7% (IQR 4·1–33·9) of maximal weight loss across operations. Weight regain was significantly greater after sleeve gastrectomy compared with RYGB: 21·3% (5·3–39·1) versus 15·0% (4·6–27·1; p<0·0001). At 5 years, RYGB resulted in significantly higher median total weight loss than sleeve gastrectomy (28·2% [21·7–35·1] vs 23·6% [15·2–31·6]) and AGB (28·2% [21·7–35·1] vs 14·9% [7·2–25·3]), both p<0·0001.

In the feature selection process during the development of the predictive weight loss model, 447 attributes were available at baseline in ABOS (appendix pp 40–60); of these, 62 (14%) were excluded because of missing data, class imbalanced, or free text input. Among the 385 remaining variables, the LASSO algorithm selected seven features that were associated with TWL at least at one postoperative visit: preoperative weight, height, type of intervention, age at intervention, current smoking history, type 2 diabetes status, and diabetes duration. The hierarchical group-LASSO method selected the same features, as well as two additional interactions: one between type of intervention and type 2 diabetes, and one between type of intervention and type 2 diabetes duration.

Training CART using the nine variables selected by hierarchical group-LASSO did not modify the decision trees. The final CART algorithm therefore used the seven features selected by LASSO for the TWL regression task.

For feature stratification, at all postoperative times, the first and most discriminant branch of decision trees divided the population by the type of intervention, with AGB being separated from RYGB at all times. At 1 year, the second descendant branch separated patients by age. The following descendant branch distinguished between sleeve gastrectomy and RYGB, only in older patients (aged >51 years), and smoking status distinguished younger patients (aged 18–51 years). At 2 years, the second descendant branch distinguished sleeve gastrectomy from RYGB. The following branch separated sleeve gastrectomy patients by age and RYGB patients according to their diabetes status. At 5 years, AGB and sleeve gastrectomy was not separated. The second descendant branch was type 2 diabetes status (RYGB patients) and age (other patients). Overall, RYGB and younger age were consistently associated with greater weight loss. Having diabetes and longer diabetes duration were always associated with less weight loss. Smoking was associated with greater weight loss, but only during the first year. The corresponding trees for each postoperative time are in the appendix (pp 25–27).

The model's prediction performances were evaluated at 1, 2, and 5 years in the testing subset of the training cohort (figure 2). At 5 years, the estimate of MAD of BMI was 3·1 kg/m² (95% CI 2·7–3·4) and RMSE of BMI was 4·9 kg/m² (3·9–5·7), corresponding to normalised estimates of 8·9% (95% CI 7·8–9·7) for MAD and 14·0% (11·2–16·3) for RMSE.

| | ABOS (n=1147) | BAREVAL (n=346) | NOK (n=5888) | SGH (n=977) | SOS (n=642) | PRECOS (n=237) | Roma (n=200) | Lyon (n=170) | COD (n=126) | Mexico (n=41) | SleevePass (n=240) | SM-BOSS (n=217) |
|---|---|---|---|---|---|---|---|---|---|---|---|---|
| Location | Lille, France | Montpellier, France | Netherlands | Singapore | Gothenburg, Sweden | Valenciennes, Arras, and Boulogne-sur-Mer, France | Roma, Italy | Lyon, France | São Paulo, Brazil | Mexico City, Mexico | Turku, Finland | Basle, Switzerland |
| Age, years | 42·1 (11·8) | 41·0 (11·7) | 44·2 (11·2) | 40·7 (10·3) | 47·3 (6·0) | 48·1 (11·6) | 44·1 (9·7) | 43·5 (10·9) | 55·7 (7·4) | 41·0 (7·6) | 48·4 (9·3) | 42·5 (11·2) |
| **Sex** | | | | | | | | | | | | |
| Female | 848 (73·9%) | 246 (71·1%) | 4683 (79·5%) | 632 (64·7%) | 450 (70·1%) | 178 (75·1%) | 122 (61·0%) | 117 (68·8%) | 66 (52·4%) | 36 (87·8%) | 167 (69·6%) | 156 (71·9%) |
| Male | 299 (26·1%) | 100 (28·9%) | 1205 (20·5%) | 345 (35·3%) | 192 (29·9%) | 59 (24·9%) | 78 (39·0%) | 53 (31·2%) | 60 (47·6%) | 5 (12·2%) | 73 (30·4%) | 61 (28·1%) |
| **Type of operation** | | | | | | | | | | | | |
| Gastric band | 223 (19·4%) | NA | 3 (0·1%) | NA | 376 (58·6%) | NA | NA | NA | NA | NA | NA | NA |
| Roux-en-Y gastric bypass | 704 (61·4%) | NA | 4801 (81·5%) | 204 (20·9%) | 266 (41·4%) | 85 (27·8%) | 165 (82·5%) | 69 (40·6%) | 126 (100·0%) | 41 (100·0%) | 119 (49·6%) | 111 (51·2%) |
| Sleeve gastrectomy | 220 (19·2%) | 346 (100·0%) | 1084 (18·4%) | 773 (79·1%) | NA | 86 (36·3%) | 35 (17·5%) | 101 (59·4%) | NA | NA | 121 (50·4%) | 106 (48·8%) |
| BMI at baseline | 47·0 (7·4) | 42·6 (5·9) | 44·2 (5·7) | 42·3 (7·3) | 42·6 (4·8) | 45·2 (7·0) | 45·2 (6·0) | 45·6 (7·6) | 34·3 (2·8) | 41·7 (5·1) | 47·9 (6·9) | 43·9 (5·3) |
| **Diabetes status at baseline** | | | | | | | | | | | | |
| No type 2 diabetes | 296 (25·8%) | 263 (76·0%) | 4528 (76·9%) | 500 (51·2%) | 357 (55·6%) | 172 (72·6%) | 111 (55·5%) | 65 (38·2%) | NA | 28 (68·3%) | 139 (57·9%) | 163 (75·1%) |
| Pre-type 2 diabetes | 454 (39·6%) | NA | NA | 102 (10·4%) | 151 (23·5%) | NA | 19 (9·5%) | 1 (0·6%) | NA | NA | NA | NA |
| Type 2 diabetes | 397 (34·6%) | 83 (24·0%) | 1360 (23·1%) | 375 (38·4%) | 134 (20·9%) | 65 (27·4%) | 70 (35·0%) | 104 (61·2%) | 126 (100·0%) | 13 (31·7%) | 101 (42·1%) | 54 (24·9%) |
| Type 2 diabetes duration in years at baseline | 21·0 (23·1) | 5·3 (6·4) | NA | 6·6 (7·7) | 3·0 (5·8) | NA | 4·1 (4·2) | 1·8 (4·3) | 8·6 (3·0) | NA | 6·3 (5·6) | 4·3 (6·0) |
| Smoking | 121 (10·5%) | 71 (20·5%) | 1115 (18·9%) | NA | 166 (25·9%) | NA | 40 (20·0%) | 12 (7·1%) | 18 (14·3%) | NA | 52 (21·7%) | NA |

Data are n (%) or mean (SD) unless stated otherwise. ABOS=Atlas Biologique de l'Obésité Sévère. BAREVAL=Medical Follow-up of Severe or Morbid Obese Patients Undergoing Bariatric Surgery. COD=Center for the treatment of Obesity and Diabetes, Hospital Oswaldo Cruz, São Paulo, Brazil. NA=not available. NOK=Nederlandse Obesitas Kliniek. PRECOS=Projet régional de REcherche Clinique en Obésité Sévère. SGH=Singapore General Hospital. SM-BOSS=Swiss Multicenter Bypass or Sleeve Study. SOS=Swedish Obese Subjects.

**Table 1: Participant characteristics at baseline in the derivation and validation cohorts**

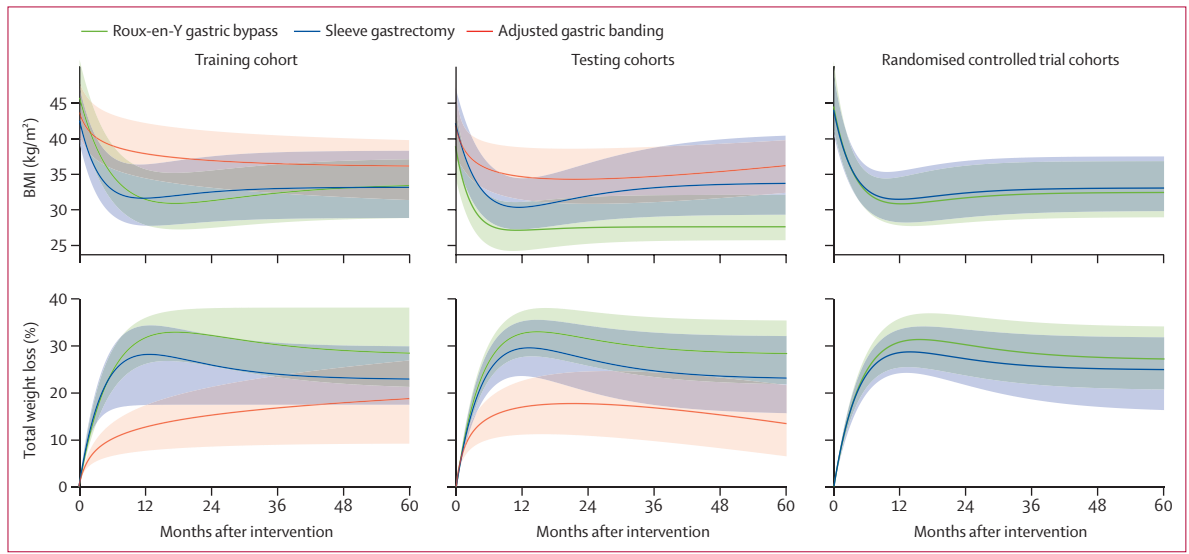

***Figure 1:*** Smoothed observed median BMI (top) and total weight loss trajectories (bottom) with corresponding IQR for each operation, for the training cohort, testing cohorts, and the randomised controlled trial cohorts

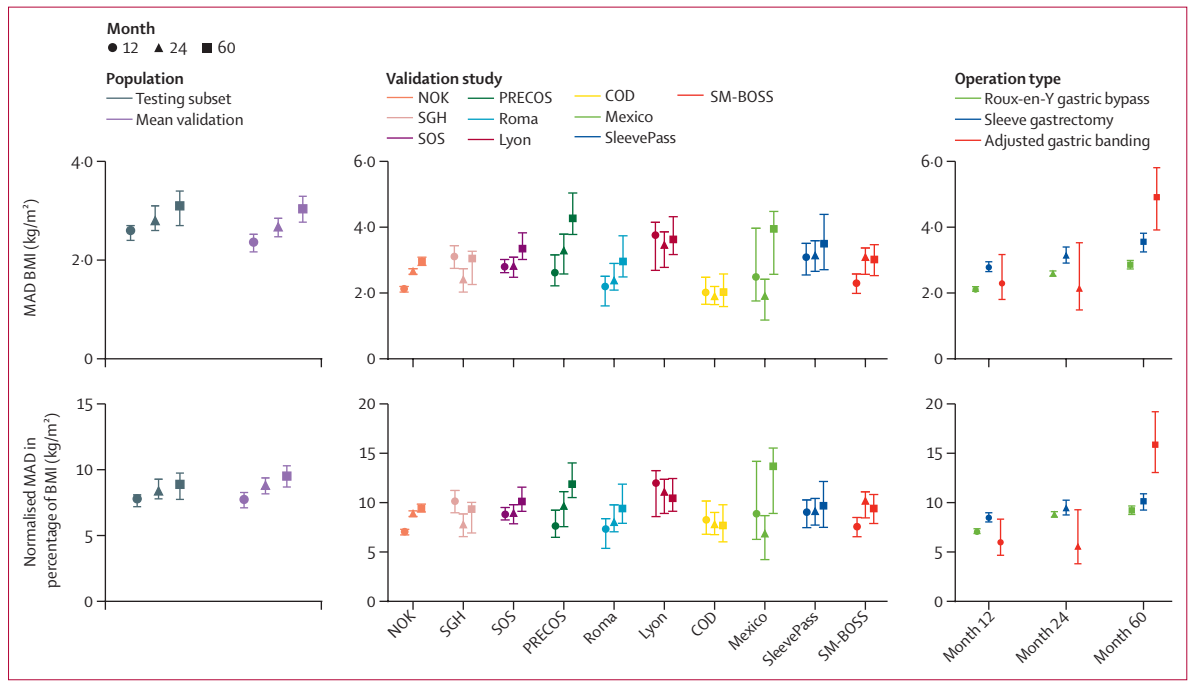

***Figure 2:*** MAD (top) and normalised MAD in percentage of BMI (bottom) of predicted outcomes in testing subset and validation cohorts, meaned by cohort size (left), individual validation cohorts (centre), and by operation (right)

COD=Center for the treatment of Obesity and Diabetes, Hospital Oswaldo Cruz, São Paulo, Brazil. MAD=median absolute deviation. NOK=Nederlandse Obesitas Kliniek. PRECOS=Projet régional de REcherche Clinique en Obésité Sévère. SGH=Singapore General Hospital. SM-BOSS=Swiss Multicenter Bypass or Sleeve Study. SOS=Swedish Obese Subjects.

The performances of the model at 1 year, 2 years, and 5 years in the testing cohorts and for each intervention are shown in figure 2, table 2, and table 3. At 5 years, across cohorts the overall mean weighted values of MAD of BMI were 2·8 kg/m² (95% CI 2·6–3·0) and RMSE of BMI were 4·7 kg/m² (4·4–5·0), corresponding to normalised estimates of 8·8% for MAD and 14·7% for

RMSE of BMI. The performances of the model were significantly higher in RYGB (MAD 2·8, RMSE 4·5) than in sleeve gastrectomy (MAD 3·5, RMSE 5·7), and AGB (MAD 4·9, RMSE 6·7), all p<0·0001. Overall, the mean difference between predicted and observed BMI at 5 years was −0·3 kg/m² (SD 4·7). The model showed good calibration at all timepoints (appendix pp 28–30),

| | BMI difference* in kg/m² (SD) | | | RMSE† in kg/m² (95% CI)‡ | | | Normalised RMSE† in percentage of BMI (95% CI)‡ | | |
|---|---|---|---|---|---|---|---|---|---|
| | Month 12 | Month 24 | Month 60 | Month 12 | Month 24 | Month 60 | Month 12 | Month 24 | Month 60 |
| Testing subset of ABOS plus BAREVAL, n=293 | −0·2 (4·1) | −0·5 (4·7) | −1·0 (4·8) | 4·1 (3·8–4·4) | 4·7 (4·0–5·4) | 4·9 (3·9–5·7) | 12·3 (11·4–13·1) | 14·0 (12·0–16·1) | 14·0 (11·2–16·3) |
| External validation | | | | | | | | | |
| NOK, n=5888 | 0·2 (3·3) | −0·1 (4·1) | −0·0 (4·7) | 3·3 (3·2–3·4) | 4·1 (4·0–4·2) | 4·7 (4·5–4·8) | 11·2 (10·8–11·3) | 13·9 (13·5–14·2) | 15·0 (14·5–15·5) |
| SGH, n=977 | −2·5 (3·7) | −0·8 (4·0) | −0·9 (4·0) | 4·4 (4·2–4·8) | 4·1 (3·7–4·6) | 4·0 (3·6–4·6) | 14·5 (13·7–15·5) | 13·2 (12·1–14·8) | 12·4 (11·0–14·2) |
| SOS, n=642 | 1·8 (4·4) | 0·9 (4·9) | −1·1 (4·9) | 4·8 (4·4–5·1) | 5·0 (4·6–5·3) | 5·0 (4·7–5·4) | 15·1 (13·9–16·1) | 15·9 (14·6–16·8) | 15·1 (13·9–16·3) |
| PRECOS, n=237 | −0·7 (4·5) | −0·0 (4·7) | 4·5 (4·0–5·0) | 4·5 (4·0–5·0) | 4·7 (4·3–5·3) | 6·2 (5·6–6·8) | 13·2 (11·8–14·7) | 13·9 (12·5–15·5) | 17·2 (15·7–19·0) |
| Roma, n=200 | −0·4 (3·8) | 1·3 (4·0) | 1·1 (4·6) | 3·8 (3·3–4·4) | 4·2 (3·6–5·2) | 4·7 (4·3–5·3) | 12·6 (10·9–14·6) | 14·1 (12·1–17·4) | 15·0 (13·6–16·8) |
| Lyon, n=170 | −1·2 (4·1) | −1·4 (4·8) | −0·5 (5·8) | 4·3 (3·7–4·9) | 5·0 (4·3–6·0) | 5·8 (5·1–6·9) | 13·5 (11·9–15·8) | 16·0 (13·7–19·4) | 16·7 (14·8–19·8) |
| COD, n=126 | −0·4 (3·1) | −1·4 (2·2) | −1·7 (2·2) | 3·1 (2·8–3·5) | 2·6 (2·3–2·9) | 2·8 (2·5–3·1) | 12·7 (11·3–14·3) | 10·6 (9·6–11·8) | 10·6 (9·6–11·7) |
| Mexico, n=41 | −0·9 (4·2) | −1·0 (3·5) | −0·3 (5·5) | 4·3 (3·4–5·5) | 3·6 (2·7–4·6) | 5·4 (4·4–7·1) | 15·4 (12·3–19·7) | 12·9 (9·9–16·4) | 18·8 (15·2–24·6) |
| SleevePass, n=240 | −1·0 (4·4) | −0·2 (4·7) | 0·2 (5·1) | 4·5 (4·1–4·9) | 4·7 (4·3–5·1) | 5·1 (4·7–5·7) | 13·2 (12·1–14·5) | 13·6 (12·4–15·0) | 14·2 (12·9–15·8) |
| SM-BOSS, n=217 | −1·1 (3·6) | 0·1 (4·4) | 0·2 (5·0) | 3·7 (3·4–4·2) | 4·4 (4·0–4·9) | 4·9 (4·5–5·7) | 12·4 (11·2–13·8) | 14·4 (13·1–16·0) | 15·5 (14·0–17·7) |
| Mean weighted by cohort sizes | −0·1 (3·5) | −0·1 (4·2) | −0·3 (4·7) | 3·7 (3·5–3·9) | 4·2 (4·0–4·5) | 4·7 (4·4–5·0) | 12·0 (11·4–12·6) | 14·0 (13·2–14·8) | 14·7 (13·8–15·7) |

ABOS=Atlas Biologique de l'Obésité Sévère. BAREVAL=Medical Follow-up of Severe or Morbid Obese Patients Undergoing Bariatric Surgery. COD=Center for the treatment of Obesity and Diabetes, Hospital Oswaldo Cruz, São Paulo, Brazil. NOK=Nederlandse Obesitas Kliniek. PRECOS=projet régional de Recherche Clinique en Obésité Sévère. RMSE=root mean squared error. SGH=Singapore General Hospital. SM-BOSS=Swiss Multicenter Bypass or Sleeve Study. SOS=Swedish Obese Subjects. *BMI difference is difference between predicted and observed BMI (negative means predicted was lower than observed). †RMSE is the measure of prediction bias and standard deviation; the lower, the more accurate. ‡95% CIs are bias-corrected and accelerated bootstrap, n=10 000 replications.

*Table 2:* Comparison of predicted outcomes in validation cohorts

| | BMI difference* in kg/m² (SD) | | | RMSE† in kg/m² (95% CI)‡ | | | Normalised RMSE† in percentage of BMI (95% CI)‡ | | |
|---|---|---|---|---|---|---|---|---|---|
| | Month 12 | Month 24 | Month 60 | Month 12 | Month 24 | Month 60 | Month 12 | Month 24 | Month 60 |
| Roux-en-Y gastric bypass | −0·0 (3·2) | −0·4 (3·9) | −0·3 (4·5) | 3·2 (3·2–3·3) | 3·9 (3·9–4·0) | 4·5 (4·3–4·6) | 11·0 (10·8–11·2) | 13·5 (13·2–13·8) | 14·6 (14·1–15·0) |
| Sleeve gastrectomy | −0·4 (4·3) | 1·0 (4·8) | 0·9 (5·6) | 4·3 (4·2–4·5) | 4·9 (4·7–5·2) | 5·7 (5·4–6·0) | 13·2 (12·7–13·8) | 14·9 (14·2–15·6) | 16·2 (15·3–17·2) |
| Adjusted gastric banding | 1·7 (3·9) | 0·7 (4·1) | −2·8 (4·3) | 4·7 (4·2–5·4) | 4·7 (4·1–5·3) | 6·0 (5·4–6·7) | 13·6 (11·9–15·3) | 13·6 (12·0–15·3) | 16·6 (14·9–18·4) |

RMSE=root mean squared error. *BMI difference is difference between predicted and observed BMI (negative means predicted was lower than observed). †RMSE is the measure of prediction bias and standard deviation; the lower, the more accurate. ‡95% CI are bias-corrected and accelerated bootstrap, n=10 000 replications.

*Table 3:* Comparison of predicted outcomes by operation in validation cohorts

that did not reveal any significant systematic bias in the predictions.

We tested the model's performance to predict the overall outcome of randomised controlled trials comparing weight loss after RYGB and sleeve gastrectomy over 5 years. The results of our synthetic study based on predicted weights were in overall agreement with those of the published research, concluding with a significantly higher weight loss following RYGB as compared to sleeve gastrectomy. The mean difference between the two groups was 14·7% (95% CI 13·7–15·7) of EWL and 6·6% (6·2–6·9) of TWL in the synthetic study versus 7·0% (3·5–10·5) of EWL and 3·2% (1·6–4·7) of TWL in the combined analysis of the two original studies.[21]

The performances of the selected decision tree model were compared in training and testing cohorts with alternative models based on simple regression, linear mixed effect, random forest, and CART with pruning. The decision tree approach outperforms simple regression at month 12, month 24, and month 60 in both internal and external test data (appendix p 32). The linear mixed model using the seven selected variables, and time as restricted cubic splines, with random intercept and slope, did not outperform the decision tree model on both internal and external test data (appendix p 33). MAD estimates are even slightly lower with the decision tree, at month 12 and month 60. Random forest resulted in only marginally lower MAD estimates, as compared with the decision tree approach (confidence intervals largely overlapped; appendix p 34). CART with pruning selected only six variables: weight, height, age, type 2 diabetes status and duration, and type of intervention, as compared to seven variables with LASSO plus CART, which additionally selected smoking history. Triaging the variables with LASSO in the first place resulted in better performances at month 12, the only timepoint where smoking history appears in the decision trees (appendix p 35).

We identified 12 models in published literature that have been previously proposed to predict weight loss following one or more of the three interventions analysed in the present study, during 1–5 years after the intervention.[26–34] The accuracy of weight loss predicted by these models, estimated by RMSE and normalised RMSE in the appropriate subset of patients from the ABOS cohort, ranged from 3·8 kg/m² to 6·1 kg/m² and from 11·7% to 17·4% of BMI at 1 year (six models), compared with means of 3·7 kg/m² and 12·0% of BMI for our current model. At 2 years, existing models ranged from 4·9 kg/m² to 7·0 kg/m² and 15·5% to 20·2% of BMI (five models), compared with 4·2 kg/m² and 14·0% of BMI for our current model. Finally, RMSE and normalised RMSE evaluated at 5 years among existing models were 5·4 kg/m² and 15·8% of BMI (only one model), compared with 4·7 kg/m² and 14·7% of BMI for our current model. Individual results of these models are in the appendix (pp 36–37).

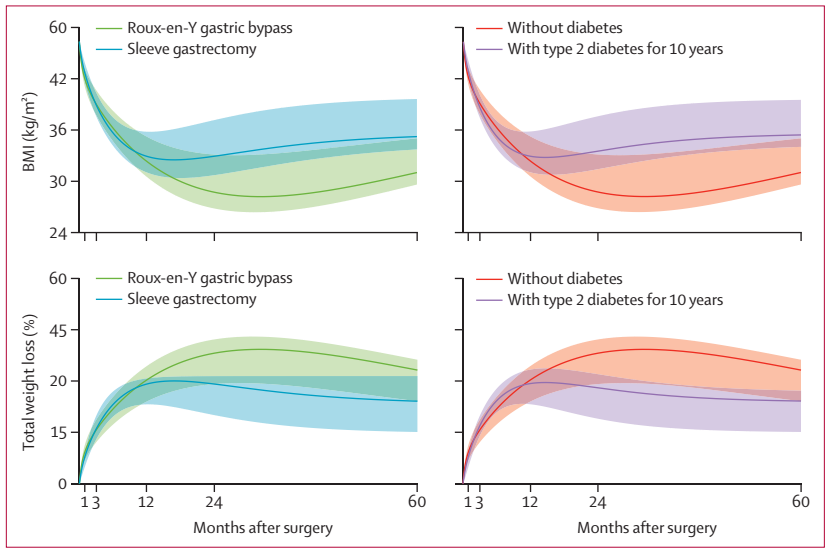

*Figure 3:* Predicted trajectory and IQR of BMI (top) and total weight loss (bottom) for a 30-year-old patient with a weight of 150 kg, a height of 1·80 m, who was a non-smoker, without diabetes, undergoing Roux-en-Y gastric bypass and sleeve gastrectomy (left); and for a 30 years old patient with a weight of 150 kg, height of 1·80 m, who was a non-smoker undergoing Roux-en-Y gastric bypass without diabetes and with type 2 diabetes with 10 years duration (right)

The machine learning model developed in the present study was then integrated into software that allowed for the display of the 5-year weight trajectory that can be expected for a given patient before the intervention according to the seven key baseline characteristics included in the model. The graphical output of the model was presented and discussed among investigators (PSau, JT, TS, MD, PP, PB, VR, HV, and FP) and patient representatives. The resulting user-friendly calculator displays the predicted weight trajectory at any given time, alongside prediction intervals corresponding to IQR of prediction errors. By default, predicted trajectories are expressed in kg. According to the user's choice, results can also be displayed in kg/m², % of TWL, or % of EWL. To improve readability, predicted trajectories over time for each of these metrics are displayed as smooth curves. Two illustrative examples of BMI and TWL trajectories predicted for individual patients from the current version of the bariatric weight trajectory prediction calculator are shown in figure 3. More examples can be found in the appendix (p 31).

## Discussion

We developed a machine learning model that provides accurate individual weight trajectories expected during 5 years after bariatric surgery, based on seven simple preoperative variables, including age, weight, height, smoking history, type 2 diabetes status and duration, and the type of intervention. These variables are readily available in a variety of clinical settings without interpretation and do not require laboratory tests. The model was validated globally, in eight cohorts and two randomised controlled trials, in Europe, the Americas, and Asia, and incorporated in an easy-to-use and interpretable web-based tool providing individual preoperative prediction of postoperative weight loss trajectory.

This accessible and interpretable model is the first to provide preoperative predictions of weight trajectories up to 5 years after surgery, simultaneously for three of the most common types of surgery: RYGB, sleeve gastrectomy, and AGB. Our results highlighted the association of the type of operation and diabetes status with weight trajectories. The present study also showed the impact of diabetes duration and smoking, which were not previously included in weight loss surgery prediction models.

As expected, the type of surgery was the first node to appear in the decision tree. Interestingly, sleeve gastrectomy and RYGB were not distinguishable at 1 year but separated soon after. This finding is consistent with literature.[35] Although some early reports showed similar weight loss after RYGB and sleeve gastrectomy at 1 year[35] and two years,[36] a meta-analysis of randomised clinical controlled trials[21] as well as a large matched controlled cohort study have shown the superiority of RYGB as compared with sleeve gastrectomy at 5 years. Additionally, a synthetic emulation of the combined analysis of the two randomised controlled trials[21] based on our model resulted in a similar conclusion to the original report, albeit resulting in a larger difference between the two operations. This finding illustrates that differences in weight loss outcomes might be larger in non-randomised settings where patient and care provider preferences drive procedure selection.[37]

Several studies have already suggested that weight loss is lower in individuals with type 2 diabetes than in individuals without diabetes particularly in patients with uncontrolled diabetes.[38–40] In the Longitudinal Assessment of Bariatric Surgery study, the presence of type 2 diabetes at baseline was associated with reduced weight loss after surgery.[41] In the present study we found that TWL also lowers with duration of diabetes increases. Diabetes duration is a predictor of disease severity, and a proxy of declined β-cell function, which was also associated with lower weight loss one year post bariatric surgery.[42]

Most of the studies that looked at preoperative smoking patterns did not show any association between smoking and postoperative weight loss.[43] A minority of studies reported small differences in postoperative weight loss between smokers and nonsmokers;[43] these differences became non-significant with longer follow-up, which is consistent with our model, which splits the population based on history of smoking only during the first year after surgery. Of note, our prediction algorithm did not identify sex as a significant predictor of postoperative weight loss, in line with several previous reports.[38,44,45]

One major strength of the present study is the use of machine learning approaches, in contrast to previous studies. LASSO is an alternative to multivariate regression that enforces sparsity in the covariates used for prediction. Notably, all variables selected by LASSO in the present study were clinical traits, as opposed to the many continuous biological variables available in ABOS. The CART algorithm learns a tree stratification of covariates that identifies relationships beyond the scope of traditional analysis techniques.[46] Tree-based models are well suited to capture non-linear effects of mixed nominal, ordinal, and continuous attributes and can also outperform deep learning on tabular data.[47] Additionally, this learned stratification of patient attributes allows for a clear interpretation of the predicted outcomes. CART also provides critical variable thresholds and their directional influences on the outcomes. Unlike multivariate linear regression approaches, CART learns non-parametric models that do not require a strong specification of the mapping between covariates and outcomes. These strengths are especially valuable in addressing the data heterogeneity commonly associated with clinical datasets. Descending in the decision tree, it is possible to draw the involvement of different features, the combination of which allows to define with higher accuracy the clinicobiological characteristics of an individual with higher or lower weight loss. Another remarkable feature is that by exploring the tree from root to leaves, CART divided the population based on diabetes status or diabetes duration, only for patients submitted to RYGB.

Our study has several limitations. First, the machine learning algorithm selected only seven simple clinical features. It is possible that the prediction can be improved with more performant classes of algorithms, such as random forest or deep learning.[48] However, such methods would require more training data and provide a less interpretable model, which we consider as a decisive criterion for implementation in patients' care settings. Indeed, more complex models are not straightforwardly interpretable by humans, undermining trust.[49] Concerns about such black-box algorithms are increasingly highlighted as one of the primary barriers to the adoption of machine learning in the health-care context.[50] Second, the extensive clinical and biological dataset used for model development only included a limited number of socioeconomic, ethnic, behavioural, and nutritional aspects, which might influence postoperative weight loss trajectories. Likewise, we did not evaluate the added value of genetic analyses[51,52] or of new disease stratification.[53] Third, our analysis was limited to the three most performed operations worldwide. AGB has been less frequently performed in the past 10 years, and several new and rapidly increasing operations, such as one anastomosis gastric bypass or endoscopic sleeve gastrectomy, were not included because of a scarcity of a large number of patients with 5 years of follow-up data. Additionally, we did not include reoperations in our model. Fourth, except for the SGH, COD, and Mexico cohorts, most individuals enrolled in our study were White, especially those who had AGB. Therefore, our results should be further replicated in non-White populations. Finally, our study was focused primarily on weight loss and not obesity complications, such as type 2 diabetes, hypertension, or non-alcoholic fatty liver disease. We also did not analyse the risks associated with surgery,[54] but we appreciate that these are also essential to inform clinical decisions.

In summary, we have developed and validated an easy-to-use and interpretable model that provides individual predictions of weight loss trajectory after bariatric surgery. We have shown its generalisability and transportability across multiple cohorts in Europe, the Americas, and Asia, as well as its performance in intervention clinical trials.

Individual weight loss trajectory prediction appears to be an accurate and simple strategy to inform clinical decisions for both health-care providers and patients before surgery. Our model can also be used postoperatively to identify patients whose actual weight loss trajectories differ from their predicted trajectory, thus allowing the timely implementation of appropriate clinical interventions.

**Contributors**
PSau, PB, VR, HV, PP, and FP conceptualised the study. PSau, PB, JT, HV, TS, RC, FP, AJ, DJ, VM, PCL, CHL, JCA-A, LC, P-AS, FG, DN, GD, MM, SA, JC, ML, EL, OV, GM, MR, PSal, RP, RVC, and CZ did data curation. PSau and PB did the statistical and mathematical analysis of the study data. PSau, PB, VR, PP, and FP developed the methodology. PSau, PB, JT, TS, and MD developed the software. PSau, PB, and JT validated the replication of the model predictions in the validation cohorts. PSau, PB, and MD visualised and developed the graphical interface and the calculator website. PSau, PB, VR, PP, and FP wrote the original draft of the manuscript. PSau, PB, VR, PP, FP, VM, and CWLR reviewed and edited the manuscript. PP and FP did project administration. FP and PP supervised the investigation and writing. CWLR acquired funding. HV, RC, FP, AJ, DJ, VM, PCL, CHL, JCA-A, LC, P-AS, FG, DN, GD, MM, SA, JC, ML, EL, OV, GM, MR, PSal, RP, RVC, CZ, and CWLR provided data. All authors have seen and approved of the final text. All authors had full access to the data and had final responsibility for the decision to submit for publication. PSau, PB, and TS directly accessed and verified the data.

**Declaration of interests**
PP reports a grant from I-Site Université Lille Nord Europe, University of Lille, Métropole Européenne de Lille, Inria, Région Hauts-de-France. RP reports a grant from Swiss National Science; and foundation fees from Johnson & Johnson and the Falik Foundation. RVC reports grants from Johnson & Johnson Medical Brazil, Medtronic Brazil, Jansen Pharmaceuticals, NovoNordisk, and Abbott; and being a member of a Scientific Advisory Board for Beineta a member of the GI Dynamics. CWLR reports grants from the Irish Research Council, Health Research Board, Science Foundation Ireland, and Anabio; being a member of the Global Advisory Board for NovoNordisk, Eli Lilly, Johnson & Johnson, Boehringer Ingelheim, GI Dynamics, Herbalife, and Irish Life Health; and has stock or stock options in Keyron and Beyond BMI. FP reports consulting fees from Novo Nordisk, Eli lilly, Medtronic, and Johnson & Johnson. All other authors declare no competing interests.

**Data sharing**
The datasets generated during or analysed during the current study are not publicly available because they are subject to national data protection laws and restrictions imposed by the ethics committee to ensure data privacy of the study participants. However, access to the datasets can be applied for through an individual project agreement with the principal investigator at the University Hospital of Lille, France (François Pattou; francois.pattou@univ-lille.fr). The authors in charge of the access and verification of the ABOS, BAREVAL, NOK, SGH, SOS, PRECOS, Roma, Lyon, COD, Mexico, SleevePass, and SM-BOSS datasets are listed in the appendix (p 65).

**Acknowledgments**
The SOPHIA Innovative Medicines Iniative (IMI) 2 Joint Undertaking under grant agreement number 875534, supported by the EU's Horizon 2020 research and innovation programme, the European Federation of Pharmaceutical Industries and Associations (EFPIA), Type 1 Diabetes Exchange, the Juvenile Diabetes Research Foundation (JDRF) and Obesity Action Coalition; Métropole Européenne de Lille; Agence Nationale de la Recherche; Institut national de recherche en sciences et technologies du numérique through the Artificial Intelligence chair Apprenf number R-PILOTE-19-004-APPRENF; Université de Lille Nord Europe's I-SITE EXPAND as part of the Bandits For Health project; Laboratoire d'excellence European Genomic Institute for Diabetes under grant ANR-10-LABX-0046; Soutien aux Travaux Interdisciplinaires, Multi-établissements et Exploratoires programme by Conseil Régional Hauts-de-France (volet partenarial phase 2, project PERSO-SURG, number 2019.01716/5). This Article is part of a project as of May 4, 2023, that has received funding from the IMI 2 Joint Undertaking under grant agreement number 875534. This Article reflects the authors view and neither IMI nor the European Union, EFPIA, or any associated partners are responsible for any use that may be made of the information contained therein.

**References**
1    Afshin A, Forouzanfar MH, Reitsma MB, et al. Health effects of overweight and obesity in 195 countries over 25 years. *N Engl J Med* 2017; **377:** 13–27.
2    Aune D, Sen A, Prasad M, et al. BMI and all cause mortality: systematic review and non-linear dose-response meta-analysis of 230 cohort studies with 3.74 million deaths among 30.3 million participants. *BMJ* 2016; **353:** i2156.
3    Carlsson LMS, Sjöholm K, Jacobson P, et al. Life expectancy after bariatric surgery in the Swedish Obese Subjects study. *N Engl J Med* 2020; **383:** 1535–43.
4    Colquitt JL, Pickett K, Loveman E, Frampton GK. Surgery for weight loss in adults. *Cochrane Database Syst Rev* 2014; **2014:** CD003641.
5    Syn NL, Cummings DE, Wang LZ, et al. Association of metabolic-bariatric surgery with long-term survival in adults with and without diabetes: a one-stage meta-analysis of matched cohort and prospective controlled studies with 174772 participants. *Lancet* 2021; **397:** 1830–41.
6    Courcoulas AP, King WC, Belle SH, et al. Seven-year weight trajectories and health outcomes in the Longitudinal Assessment of Bariatric Surgery (LABS) study. *JAMA Surg* 2018; **153:** 427–34.
7    Peterli R, Wölnerhanssen BK, Peters T, et al. Effect of laparoscopic sleeve gastrectomy vs laparoscopic Roux-en-Y gastric bypass on weight loss in patients with morbid obesity: the SM-BOSS randomized clinical trial. *JAMA* 2018; **319:** 255–65.
8    Karpińska IA, Kulawik J, Pisarska-Adamczyk M, Wysocki M, Pędziwiatr M, Major P. Is it possible to predict weight loss after bariatric surgery–external validation of predictive models. *Obes Surg* 2021; **31:** 2994–3004.
9    Puzziferri N, Roshek TB 3rd, Mayo HG, Gallagher R, Belle SH, Livingston EH. Long-term follow-up after bariatric surgery: a systematic review. *JAMA* 2014; **312:** 934–42.
10   Tettero OM, Monpellier VM, Janssen IMC, Steenhuis IHM, van Stralen MM. Early postoperative weight loss predicts weight loss up to 5 years after Roux-en-Y gastric bypass, banded Roux-en-Y gastric bypass, and sleeve gastrectomy. *Obes Surg* 2022; **32:** 2891–902.
11   Manning S, Pucci A, Carter NC, et al. Early postoperative weight loss predicts maximal weight loss after sleeve gastrectomy and Roux-en-Y gastric bypass. *Surg Endosc* 2015; **29:** 1484–91.
12   Batterham M, Tapsell LC, Charlton KE. Predicting dropout in dietary weight loss trials using demographic and early weight change characteristics: implications for trial design. *Obes Res Clin Pract* 2016; **10:** 189–96.
13   Finks JF, English WJ, Carlin AM, et al. Predicting risk for venous thromboembolism with bariatric surgery: results from the Michigan Bariatric Surgery Collaborative. *Ann Surg* 2012; **255:** 1100–04.
14   Bertsimas D, Dunn J, Velmahos GC, Kaafarani HMA. Surgical risk is not linear: derivation and validation of a novel, user-friendly, and machine-learning-based predictive optimal trees in emergency surgery risk (POTTER) calculator. *Ann Surg* 2018; **268:** 574–83.
15   Bektaş M, Reiber BMM, Pereira JC, Burchell GL, van der Peet DL. Artificial intelligence in bariatric surgery: current status and future perspectives. *Obes Surg* 2022; **32:** 2772–83.
16   Mingrone G, Panunzi S, De Gaetano A, et al. Metabolic surgery versus conventional medical therapy in patients with type 2 diabetes: 10-year follow-up of an open-label, single-centre, randomised controlled trial. *Lancet* 2021; **397:** 293–304.
17   Tan SYT, Syn NL, Lin DJ, et al. Centile charts for monitoring of weight loss trajectories after bariatric surgery in Asian patients. *Obes Surg* 2021; **31:** 4781–89.
18   Cohen RV, Pereira TV, Aboud CM, et al. Effect of gastric bypass vs best medical treatment on early-stage chronic kidney disease in patients with type 2 diabetes and obesity: a randomized clinical trial. *JAMA Surg* 2020; **155:** e200420.
19   Zerrweck C, Herrera A, Sepúlveda EM, Rodríguez FM, Guilbert L. Long versus short biliopancreatic limb in Roux-en-Y gastric bypass: short-term results of a randomized clinical trial. *Surg Obes Relat Dis* 2021; **17:** 1425–30.
20   Salminen P, Grönroos S, Helmiö M, et al. Effect of laparoscopic sleeve gastrectomy vs Roux-en-Y gastric bypass on weight loss, comorbidities, and reflux at 10 years in adult patients with obesity: the SLEEVEPASS randomized clinical trial. *JAMA Surg* 2022; **157:** 656–66.
21   Wölnerhanssen BK, Peterli R, Hurme S, et al. Laparoscopic Roux-en-Y gastric bypass versus laparoscopic sleeve gastrectomy: 5-year outcomes of merged data from two randomized clinical trials (SLEEVEPASS and SM-BOSS). *Br J Surg* 2021; **108:** 49–57.
22   Lim M, Hastie T. Learning interactions via hierarchical group-lasso regularization. *J Comput Graph Stat* 2015; **24:** 627–54.
23   Breiman L, Friedman J, Stone CJ, Olshen RA. Classification and regression trees. New York, NY: Taylor & Francis, 1984.
24   Salminen P, Helmiö M, Ovaska J, et al. Effect of laparoscopic sleeve gastrectomy vs laparoscopic Roux-en-Y gastric bypass on weight loss at 5 years among patients with morbid obesity: the SLEEVEPASS randomized clinical trial. *JAMA* 2018; **319:** 241–54.
25   Van Buuren S, Groothuis-Oudshoorn K. mice: multivariate imputation by chained equations in R. *J Stat Softw* 2011; **45:** 1–67.
26   Therneau T, Atkinson B, Ripley B, Ripley MB. Package 'rpart'. 2015. https://cran.r-project.org/web/packages/rpart/rpart.pdf (accessed Oct 10, 2022).
27   Collins GS, Dhiman P, Andaur Navarro CL, et al. Protocol for development of a reporting guideline (TRIPOD-AI) and risk of bias tool (PROBAST-AI) for diagnostic and prognostic prediction model studies based on artificial intelligence. *BMJ Open* 2021; **11:** e048008.
28   Baltasar A, Perez N, Serra C, Bou R, Bengochea M, Borrás F. Weight loss reporting: predicted body mass index after bariatric surgery. *Obes Surg* 2011; **21:** 367–72.
29   Wise ES, Hocking KM, Kavic SM. Prediction of excess weight loss after laparoscopic Roux-en-Y gastric bypass: data from an artificial neural network. *Surg Endosc* 2016; **30:** 480–88.

30  Goulart A, Leão P, Costa P, et al. Doctor, how much weight will I lose?–a new individualized predictive model for weight loss. *Obes Surg* 2016; **26:** 1357–59.

31  Seyssel K, Suter M, Pattou F, et al. A predictive model of weight loss after Roux-en-Y gastric bypass up to 5 years after surgery: a useful tool to select and manage candidates for bariatric surgery. *Obes Surg* 2018; **28:** 3393–99.

32  Janik MR, Rogula TG, Mustafa RR, Saleh AA, Abbas M, Khaitan L. Setting realistic expectations for weight loss after laparoscopic sleeve gastrectomy. *Wideochir Inne Tech Malo Inwazyjne* 2019; **14:** 415–19.

33  Velázquez-Fernández D, Sánchez H, Monraz F, et al. Development of an interactive outcome estimation tool for laparoscopic Roux-en-Y gastric bypass in Mexico based on a cohort of 1002 patients. *Obes Surg* 2019; **29:** 2878–85.

34  Cottam S, Cottam D, Cottam A, Zaveri H, Surve A, Richards C. The use of predictive markers for the development of a model to predict weight loss following vertical sleeve gastrectomy. *Obes Surg* 2018; **28:** 3769–74.

35  Peterli R, Steinert RE, Woelnerhanssen B, et al. Metabolic and hormonal changes after laparoscopic Roux-en-Y gastric bypass and sleeve gastrectomy: a randomized, prospective trial. *Obes Surg* 2012; **22:** 740–48.

36  Fischer L, Hildebrandt C, Bruckner T, et al. Excessive weight loss after sleeve gastrectomy: a systematic review. *Obes Surg* 2012; **22:** 721–31.

37  Arterburn DE, Johnson E, Coleman KJ, et al. Weight outcomes of sleeve gastrectomy and gastric bypass compared to nonsurgical treatment. *Ann Surg* 2021; **274:** e1269–76.

38  Parri A, Benaiges D, Schröder H, et al. Preoperative predictors of weight loss at 4 years following bariatric surgery. *Nutr Clin Pract* 2015; **30:** 420–24.

39  Shantavasinkul PC, Omotosho P, Corsino L, Portenier D, Torquati A. Predictors of weight regain in patients who underwent Roux-en-Y gastric bypass surgery. *Surg Obes Relat Dis* 2016; **12:** 1640–45.

40  Diedisheim M, Poitou C, Genser L, et al. Weight loss after sleeve gastrectomy: does type 2 diabetes status impact weight and body composition trajectories? *Obes Surg* 2021; **31:** 1046–54.

41  Courcoulas AP, Christian NJ, O'Rourke RW, et al. Preoperative factors and 3-year weight change in the Longitudinal Assessment of Bariatric Surgery (LABS) consortium. *Surg Obes Relat Dis* 2015; **11:** 1109–18.

42  Borges-Canha M, Neves JS, Mendonça P, et al. Beta cell function as a baseline predictor of weight loss after bariatric surgery. *Front Endocrinol (Lausanne)* 2021; **12:** 714173.

43  Mohan S, Samaan JS, Samakar K. Impact of smoking on weight loss outcomes after bariatric surgery: a literature review. *Surg Endosc* 2021; **35:** 5936–52.

44  Mousapour P, Tasdighi E, Khalaj A, et al. Sex disparity in laparoscopic bariatric surgery outcomes: a matched-pair cohort analysis. *Sci Rep* 2021; **11:** 12809.

45  Tankel J, Shlezinger O, Neuman M, et al. Predicting weight loss and comorbidity improvement 7 years following laparoscopic sleeve gastrectomy: does early weight loss matter? *Obes Surg* 2020; **30:** 2505–10.

46  Barnholtz-Sloan JS, Guan X, Zeigler-Johnson C, Meropol NJ, Rebbeck TR. Decision tree-based modeling of androgen pathway genes and prostate cancer risk. *Cancer Epidemiol Biomarkers Prev* 2011; **20:** 1146–55.

47  Grinsztajn L, Oyallon E, Varoquaux G. Why do tree-based models still outperform deep learning on typical tabular data? *Adv Neural Inf Process Syst* 2022; **35:** 507–20.

48  Ge G, Wong GW. Classification of premalignant pancreatic cancer mass-spectrometry data using decision tree ensembles. *BMC Bioinformatics* 2008; **9:** 275.

49  Stiglic G, Kocbek P, Fijacko N, Zitnik M, Verbert K, Cilar L. Interpretability of machine learning-based prediction models in healthcare. *Wiley Interdiscip Rev Data Min Knowl Discov* 2020; **10:** e1379.

50  Petch J, Di S, Nelson W. Opening the black box: the promise and limitations of explainable machine learning in cardiology. *Can J Cardiol* 2022; **38:** 204–13.

51  de Toro-Martin J, Guénard F, Tchernof A, Pérusse L, Marceau S, Vohl MC. Polygenic risk score for predicting weight loss after bariatric surgery. *JCI Insight* 2018; **3:** e122011.

52  Antoine D, Guéant-Rodriguez RM, Chèvre JC, et al. Low-frequency coding variants associated with body mass index affect the success of bariatric surgery. *J Clin Endocrinol Metab* 2022; **107:** e1074–84.

53  Raverdy V, Cohen RV, Caiazzo R, et al. Data-driven subgroups of type 2 diabetes, metabolic response, and renal risk profile after bariatric surgery: a retrospective cohort study. *Lancet Diabetes Endocrinol* 2022; **10:** 67–76.

54  Thereaux J, Lesuffleur T, Czernichow S, et al. Long-term adverse events after sleeve gastrectomy or gastric bypass: a 7-year nationwide, observational, population-based, cohort study. *Lancet Diabetes Endocrinol* 2019; **7:** 786–95.